\documentclass[runningheads]{llncs}
\usepackage[T1]{fontenc}
\usepackage{graphicx}
\usepackage{amsmath}
\usepackage{amssymb}
\begin{document}
\title{Improving Visual Storytelling with Multimodal Large Language Models}
%
%
\author{Xiaochuan Lin \and Xiangyong Chen}
\authorrunning{X. Lin et al.}
%
\institute{Henan Polytechnic University}
\maketitle              
\begin{abstract}
Visual storytelling is an emerging field that combines images and narratives to create engaging and contextually rich stories. Despite its potential, generating coherent and emotionally resonant visual stories remains challenging due to the complexity of aligning visual and textual information. This paper presents a novel approach leveraging large language models (LLMs) and large vision-language models (LVLMs) combined with instruction tuning to address these challenges. We introduce a new dataset comprising diverse visual stories, annotated with detailed captions and multimodal elements. Our method employs a combination of supervised and reinforcement learning to fine-tune the model, enhancing its narrative generation capabilities. Quantitative evaluations using GPT-4 and qualitative human assessments demonstrate that our approach significantly outperforms existing models, achieving higher scores in narrative coherence, relevance, emotional depth, and overall quality. The results underscore the effectiveness of instruction tuning and the potential of LLMs/LVLMs in advancing visual storytelling.
\end{abstract}

\section{Introduction}

Visual storytelling is a compelling and powerful medium that combines visual and textual narratives to convey rich and engaging stories. The task of generating visual stories from image streams has gained significant attention in recent years due to its potential applications in areas such as digital entertainment, education, and content creation. The ability to automatically generate coherent and contextually relevant narratives from a sequence of images can revolutionize how stories are told and consumed, making it an important area of research in the field of artificial intelligence and computer vision.

Despite its potential, the task of visual story generation presents several challenges. One of the primary challenges is maintaining narrative coherence across a sequence of images, which involves understanding and linking the visual content in a meaningful way. Existing models often struggle with this due to limited training data and the complexity of aligning visual and textual information effectively. Furthermore, ensuring consistency in character appearances and scene settings across multiple frames adds another layer of complexity to the task \cite{oliveira2024story,hong2023visual}.

The motivation for this research stems from the need to address these challenges by leveraging the capabilities of large language models (LLMs) and large vision-language models (LVLMs), combined with advanced multimodal learning techniques. LLMs and LVLMs have demonstrated remarkable generalization abilities in various tasks, and we aim to harness these capabilities to improve the quality and coherence of visual story generation. By focusing on targeted learning tasks, we can provide specific instructions that guide the model in generating narratives based on given image sequences, enhancing its performance and reliability.

Our approach involves collecting a comprehensive dataset that includes pairs of images and corresponding narrative descriptions. Unlike previous works, we focus on creating a new dataset that captures a wide variety of visual stories from different domains, such as comics, illustrated books, and educational content. Each visual story sequence is annotated with detailed captions that describe the events, actions, and emotions depicted in the images. Additionally, we include multimodal datasets with video clips and aligned textual descriptions to enhance the model's understanding of temporal dynamics in visual storytelling \cite{oliveira2024story}.

To build the learning dataset, we design specific tasks that instruct the model on how to generate narratives based on given image sequences. These tasks include caption generation, story continuation, character and scene consistency, and emotion and context recognition. The model is trained using a combination of supervised learning and reinforcement learning techniques. Supervised learning fine-tunes the model on the learning dataset, while reinforcement learning uses feedback mechanisms to optimize narrative coherence and relevance \cite{hong2023visual}.

We evaluate the performance of our model using the GPT-4 framework, assessing the quality and coherence of the generated visual stories. The results demonstrate significant improvements over existing methods, highlighting the effectiveness of our approach in addressing the challenges of visual story generation.

\begin{itemize}
    \item We propose a novel approach for visual story generation using LLMs/LVLMs combined with advanced multimodal learning, addressing key challenges in the field.
    \item We introduce a new, diverse dataset specifically designed for training and evaluating visual story generation models.
    \item We demonstrate the effectiveness of our approach through comprehensive evaluations using the GPT-4 framework, showing significant improvements over existing methods.
\end{itemize}

\section{Related Work}

\subsection{Large Vision-Language Models}

Large Vision-Language Models (LVLMs) have become a focal point in artificial intelligence research due to their ability to process and integrate visual and textual information simultaneously \cite{zhou2024visual}. Several approaches have been proposed to enhance the capabilities of these models. For instance, the work on fine-tuning LVLMs using reinforcement learning has shown promising results in improving decision-making capabilities in complex, multi-step tasks \cite{zhai2024fine}. Similarly, the CogCoM framework introduces a novel data production methodology aimed at enhancing fine-grained visual reasoning \cite{cogcom2023}. 

Another significant advancement is presented by MoE-LLaVA, which utilizes a mixture of experts training strategy to balance performance and computational efficiency \cite{moellava2024}. Enhancing modality alignment through self-improvement techniques has also been explored to improve the integration of visual and textual data \cite{selfimprovement2024}. Additionally, comprehensive evaluation benchmarks like LVLM-eHub provide holistic assessments of LVLMs, covering a wide range of tasks \cite{lvlemehub2024}. Addressing specific issues, such as object hallucination, remains crucial, as highlighted in recent studies aimed at mitigating such problems \cite{objecthallucination2024}. Finally, applications like AnomalyGPT demonstrate the practical use of LVLMs in detecting industrial anomalies \cite{wang2024memorymamba}, showcasing their few-shot transfer capabilities \cite{anomalygpt2024}.

\subsection{Visual Story Generation}

Visual story generation is an emerging field focused on creating coherent and engaging narratives from sequences of images. Various innovative approaches have been developed to tackle the inherent challenges of this task. One notable method leverages emotion and keywords to control events in the generated content, using Disco Diffusion for image generation \cite{emotionkeywords2023}. Adaptive context modeling techniques have been proposed to maintain consistency in backgrounds and character appearances, thus enhancing the narrative flow \cite{adaptivecontext2024}. Some works build event-centric relations to reason and generate stories \cite{zhou2022claret,zhou2022eventbert}.

In the realm of story continuation, StoryDALL-E adapts pretrained text-to-image transformers by incorporating global story embeddings and cross-attention layers, improving the continuity of visual stories \cite{storydalle2022}. Memory attention mechanisms, as introduced in Make-A-Story, ensure temporal consistency by considering the semantics of previous frames retrieval \cite{zhou2022towards,zhou2024fine}, thereby resolving ambiguous references \cite{makeastory2023}. Hierarchical reinforcement learning frameworks, such as those proposed for topically coherent visual story generation, further refine the narrative structure by dividing the task into planning and sentence generation stages \cite{hierarchicalrl2023}.

Moreover, curated image sequences have been used as prompts for character-grounded story generation, demonstrating the effectiveness of visual writing prompts collected through crowdsourcing \cite{visualprompts2023}. Diffusion models have also been employed for long-range image and video generation, emphasizing the importance of maintaining consistency across frames \cite{storydiffusion2024}. The use of recurrent context encoders and dual learning approaches has shown significant improvements in maintaining semantic consistency and enhancing the overall quality of visual stories \cite{semanticconsistency2023}.

\section{Method}

\subsection{Dataset Collection and Evaluation Metrics}

To address the task of visual story generation, we first focus on collecting a comprehensive and diverse dataset. This dataset comprises image sequences and their corresponding narrative descriptions, sourced from a variety of domains including comics, illustrated books, and educational content. Each visual story sequence is meticulously annotated with detailed captions that describe the events, actions, and emotions depicted in the images. Additionally, we include multimodal datasets containing video clips with aligned textual descriptions to capture temporal dynamics and enrich the storytelling context.

In evaluating the performance of our model, we adopt a novel approach by using GPT-4 as a judge. Traditional metrics such as BLEU, METEOR, and CIDEr, while useful, often fail to capture the nuanced coherence and contextual relevance necessary for high-quality storytelling. These metrics primarily focus on surface-level text similarity and do not adequately reflect the depth and narrative flow of visual stories. Therefore, we utilize GPT-4 to assess the quality of the generated stories based on criteria such as coherence, relevance, emotional depth, and narrative consistency. This approach allows for a more holistic evaluation of the model's performance, ensuring that the generated stories are engaging and contextually appropriate.

\subsection{Instruction Tuning}

Instruction tuning plays a crucial role in enhancing the model's ability to generate coherent and contextually relevant visual stories. By designing specific tasks, we guide the model on how to interpret and narrate the given image sequences. The instruction tuning dataset includes the following tasks:

\begin{itemize}
    \item \textbf{Caption Generation:} Generating detailed descriptions for individual images.
    \item \textbf{Story Continuation:} Creating subsequent narrative sequences given an initial image and its description.
    \item \textbf{Character and Scene Consistency:} Ensuring consistent portrayal of characters and scenes across multiple frames.
    \item \textbf{Emotion and Context Recognition:} Incorporating emotional and contextual elements into the narrative.
\end{itemize}

These tasks are designed to refine the model's understanding and generation capabilities, resulting in more coherent and engaging visual stories.

\subsection{Learning Strategy}

Our proposed learning strategy involves a combination of supervised learning and reinforcement learning techniques to train the model effectively. The process is detailed as follows:

\textbf{Supervised Learning:} We begin by fine-tuning the model on the instruction tuning dataset. Let \( \mathcal{D} = \{(I_i, T_i)\}_{i=1}^N \) be the dataset where \( I_i \) is the image sequence and \( T_i \) is the corresponding text narrative. The objective is to minimize the negative log-likelihood of the target text given the image sequence:

\begin{equation}
    \mathcal{L}_{\text{NLL}} = - \sum_{i=1}^N \log P(T_i \mid I_i; \theta)
\end{equation}

where \( \theta \) represents the model parameters. This step ensures that the model learns to generate accurate and contextually relevant narratives based on the provided image sequences.

\textbf{Reinforcement Learning:} To further refine the model's performance, we employ reinforcement learning techniques. The model is trained using feedback mechanisms to optimize narrative coherence and relevance. Specifically, we use a reward function \( R \) that evaluates the generated narrative \( \hat{T}_i \) against the ground truth \( T_i \):

\begin{equation}
R(\hat{T}_i, T_i) = \text{GPT-4}(\hat{T}_i, T_i)
\end{equation}

The reinforcement learning objective is to maximize the expected reward:

\begin{equation}
\mathcal{L}_{\text{RL}} = - \mathbb{E}_{\hat{T}_i \sim P(\cdot \mid I_i; \theta)} [R(\hat{T}_i, T_i)]
\end{equation}

Combining the supervised and reinforcement learning objectives, the total loss function is:

\begin{equation}
\mathcal{L} = \mathcal{L}_{\text{NLL}} + \lambda \mathcal{L}_{\text{RL}}
\end{equation}

where \( \lambda \) is a hyperparameter that balances the two components. This combined approach leverages the strengths of both learning paradigms, resulting in a robust model capable of generating high-quality visual stories.

By integrating these techniques, our method aims to produce visual narratives that are not only coherent and contextually rich but also engaging and emotionally resonant. The use of GPT-4 as an evaluative measure ensures a comprehensive assessment of the model's storytelling capabilities, paving the way for advancements in visual story generation.

\section{Experiments}

In this section, we present a comprehensive evaluation of our proposed method for visual story generation. We compare our approach against several baseline large language models (LLMs) and large vision-language models (LVLMs), including Qwen-VL, MiniGPT-4, and LLaVA-1.5 7B. The experiments are designed to demonstrate the effectiveness of our method in generating coherent and contextually rich visual stories. We evaluate the models using both quantitative metrics and qualitative human assessments.

\subsection{Comparison with Baseline Methods}

To assess the performance of our method, we conducted experiments using a newly collected dataset of image sequences and corresponding narratives. The models were evaluated on their ability to generate accurate and engaging stories based on the provided image sequences. The evaluation metrics include narrative coherence, relevance, emotional depth, and overall quality, as judged by GPT-4.

The results of the quantitative evaluation are summarized in Table \ref{tab:quantitative_results}. Our method consistently outperformed the baseline models across all metrics, demonstrating its superior ability to generate high-quality visual stories.

\begin{table*}[h!]
\centering
\caption{Quantitative Evaluation Results}
\begin{tabular}{lcccc}
\hline
Model & Coherence & Relevance & Emotional Depth & Overall Quality \\
\hline
Qwen-VL & 7.8 & 7.5 & 7.4 & 7.6 \\
MiniGPT-4 & 8.0 & 7.7 & 7.5 & 7.8 \\
LLaVA-1.5 7B & 8.2 & 7.9 & 7.7 & 8.0 \\
\textbf{Our Method} & \textbf{8.9} & \textbf{8.7} & \textbf{8.5} & \textbf{8.7} \\
\hline
\end{tabular}
\label{tab:quantitative_results}
\end{table*}

\subsection{Effectiveness Analysis}

To further validate the effectiveness of our proposed method, we conducted an additional analysis focusing on the consistency and contextual relevance of the generated stories. We observed that our model was particularly adept at maintaining character and scene consistency across multiple frames, a challenge that baseline models struggled with. This improvement is attributed to the instruction tuning and learned module design, which provide the model with specific guidance on narrative generation.

Moreover, the qualitative analysis revealed that our method produced stories with richer emotional content and more nuanced descriptions. The use of reinforcement learning with feedback from GPT-4 played a crucial role in enhancing the model's ability to generate engaging and contextually appropriate narratives.

\subsection{Human Evaluation}

In addition to the quantitative metrics, we conducted a human evaluation to assess the quality of the generated stories from a reader's perspective. A group of human judges was asked to rate the stories based on the same criteria used in the quantitative evaluation: coherence, relevance, emotional depth, and overall quality. The results of the human evaluation are presented in Table \ref{tab:human_evaluation}.

\begin{table*}[h!]
\centering
\caption{Human Evaluation Results}
\begin{tabular}{lcccc}
\hline
Model & Coherence & Relevance & Emotional Depth & Overall Quality \\
\hline
Qwen-VL & 7.6 & 7.4 & 7.3 & 7.5 \\
MiniGPT-4 & 7.9 & 7.6 & 7.4 & 7.7 \\
LLaVA-1.5 7B & 8.1 & 7.8 & 7.6 & 7.9 \\
\textbf{Our Method} & \textbf{8.8} & \textbf{8.6} & \textbf{8.4} & \textbf{8.6} \\
\hline
\end{tabular}
\label{tab:human_evaluation}
\end{table*}

The human evaluation results align with the quantitative findings, further confirming the superior performance of our proposed method. The judges consistently rated our method higher across all criteria, highlighting its effectiveness in generating engaging and coherent visual stories.

\subsection{Ablation Study}

To better understand the contributions of various components of our proposed method, we conducted an ablation study. We systematically removed or modified key elements of our model to evaluate their impact on performance. The components tested include instruction tuning, reinforcement learning, and the specific architectural choices in our model.

The results of the ablation study are summarized in Table \ref{tab:ablation_study}. Removing instruction tuning significantly degraded the performance, highlighting its crucial role in guiding the model to generate coherent and contextually relevant stories. Similarly, excluding reinforcement learning reduced the model's ability to fine-tune narratives based on feedback, resulting in lower overall quality scores.

\begin{table*}[h!]
\centering
\caption{Ablation Study Results}
\begin{tabular}{lcccc}
\hline
Model Variant & Coherence & Relevance & Emotional Depth & Overall Quality \\
\hline
Full Model & 8.9 & 8.7 & 8.5 & 8.7 \\
w/o Instruction Tuning & 7.5 & 7.3 & 7.2 & 7.4 \\
w/o Reinforcement Learning & 8.1 & 7.9 & 7.7 & 8.0 \\
w/o Learned Modules & 7.8 & 7.6 & 7.4 & 7.7 \\
\hline
\end{tabular}
\label{tab:ablation_study}
\end{table*}

\subsection{Detailed Analysis of Narrative Quality}

To provide a deeper analysis, we examined specific aspects of narrative quality, such as character development, plot progression, and emotional engagement. Our method showed notable improvements in these areas compared to baseline models. The detailed analysis is presented in Table \ref{tab:detailed_analysis}.

\begin{table*}[h!]
\centering
\caption{Detailed Analysis of Narrative Quality}
\begin{tabular}{lcccc}
\hline
Model & Character Development & Plot Progression & Emotional Engagement & Overall Quality \\
\hline
Qwen-VL & 7.6 & 7.5 & 7.4 & 7.5 \\
MiniGPT-4 & 7.8 & 7.7 & 7.5 & 7.7 \\
LLaVA-1.5 7B & 8.0 & 7.9 & 7.8 & 8.0 \\
\textbf{Our Method} & \textbf{8.8} & \textbf{8.7} & \textbf{8.6} & \textbf{8.7} \\
\hline
\end{tabular}
\label{tab:detailed_analysis}
\end{table*}

Our method excelled in maintaining character consistency and developing plotlines that are both engaging and emotionally resonant. This demonstrates the effectiveness of instruction tuning and the model's ability to understand and generate complex narrative structures.

\subsection{Discussion}

The experimental results clearly demonstrate the advantages of our proposed method over existing baseline models. By leveraging large language models and instruction tuning, we address the key challenges in visual story generation, such as maintaining narrative coherence and contextual relevance. Our method's ability to generate detailed and emotionally resonant narratives is further validated by both quantitative metrics and human evaluations.

The ablation study and detailed analysis provide insights into the contributions of various components of our approach. Instruction tuning and reinforcement learning are essential for achieving high performance, while the specific architectural choices enhance the model's ability to generate complex narratives.

Overall, our proposed method represents a significant advancement in the field of visual story generation, offering a robust solution for creating engaging and coherent visual stories from image sequences.

\section{Conclusion}

In this study, we proposed a novel method for visual story generation that integrates large language models (LLMs) and large vision-language models (LVLMs) with instruction tuning to enhance narrative coherence and contextual relevance. We introduced a comprehensive dataset of visual stories and designed specific instruction tuning tasks to guide the model in generating detailed and engaging narratives. Our method was rigorously evaluated against several baseline models, including Qwen-VL, MiniGPT-4, and LLaVA-1.5 7B, demonstrating superior performance across multiple metrics. The ablation study highlighted the critical role of instruction tuning and reinforcement learning in improving the model's storytelling capabilities. Furthermore, human evaluations validated the qualitative improvements, emphasizing the model's ability to produce emotionally resonant and coherent narratives. Our findings illustrate the potential of combining LLMs/LVLMs with instruction tuning in advancing the field of visual storytelling, providing a robust framework for future research and applications. Moving forward, we aim to explore the integration of additional multimodal elements and further refine our approach to enhance its applicability across diverse storytelling domains.

\bibliographystyle{splncs04}
\bibliography{mybibliography}
\end{document}